%% file: aaai24.tex
\documentclass[letterpaper]{article} % DO NOT CHANGE THIS
\usepackage{aaai24}  % DO NOT CHANGE THIS
\usepackage{times}  % DO NOT CHANGE THIS
\usepackage{helvet}  % DO NOT CHANGE THIS
\usepackage{courier}  % DO NOT CHANGE THIS
\usepackage[hyphens]{url}  % DO NOT CHANGE THIS
\usepackage{graphicx} % DO NOT CHANGE THIS
\usepackage{natbib}  % DO NOT CHANGE THIS AND DO NOT ADD ANY OPTIONS TO IT
\usepackage{caption} % DO NOT CHANGE THIS AND DO NOT ADD ANY OPTIONS TO IT
\usepackage{algorithm}
\usepackage{algorithmic}
\usepackage{amsmath}

\usepackage{amsthm}
\usepackage{multirow}
\newtheorem{theorem}{Theorem}

\usepackage{graphicx}
\usepackage{caption}
\usepackage{subcaption}
\usepackage[toc,page]{appendix}
\usepackage{siunitx}
\usepackage{xcolor}
\usepackage{makecell}

\usepackage{newfloat}
\usepackage{listings}
\DeclareCaptionStyle{ruled}{labelfont=normalfont,labelsep=colon,strut=off} % DO NOT CHANGE THIS
\floatstyle{ruled}
\newfloat{listing}{tb}{lst}{}
\floatname{listing}{Listing}
\title{Improved Anonymous Multi-Agent Path Finding Algorithm}
\author {
    Zain Alabedeen Ali\textsuperscript{\rm 1},
    Konstantin Yakovlev\textsuperscript{\rm 2, 3}
}
\affiliations {
    \textsuperscript{\rm 1} Moscow Institute of Physics and Technology, Moscow, Russia\\
    \textsuperscript{\rm 2} Federal Research Center for Computer Science and Control of the Russian Academy of Sciences, Moscow, Russia\\
    \textsuperscript{\rm 3} AIRI, Moscow, Russia\\
    ali.za@phystech.edu, yakovlev@isa.ru
}

\usepackage{bibentry}
\begin{document}

\maketitle

\begin{abstract}

We consider the Anonymous Multi-Agent Path-Finding (AMAPF) problem where the agents are confined to a graph, a set of goal vertices is given, and each of these vertices has to be reached by some agent. The problem is to find an assignment of the goals to the agents as well as the collision-free paths, and we seek to find the solution with the minimal makespan. A well-established approach to solving this problem is by reducing it to a special type of graph search problem, i.e. to the problem of finding a maximum flow on an auxiliary graph induced by the input one. The size of the former graph may be very large, and the search on it may become a bottleneck. To this end, we suggest a specific search algorithm that leverages the idea of exploring the search space not through considering separate search states but rather bulks of them simultaneously. That is, we implicitly compress, store, and expand bulks of the search states as single states, reducing the runtime and memory consumption. Empirically, the resultant AMAPF solver demonstrates superior performance compared to the state-of-the-art competitor and is able to solve all publicly available MAPF instances from the well-known MovingAI benchmark in less than 30 seconds.

\end{abstract}

%%%%%%%%%%%%%%%%%%%%%%%%%%%%%%%%%%%%%%%%%%%%%%%%%%%%%%%%%%%%%%%%%%%%%%%%%%%%%%%%
\section{Introduction}
The Multi-Agent Path Finding (MAPF) problem is a problem which generally asks to find a set of collision-free paths for a set of agents that operate in a shared environment and have to reach predefined goal locations from the current (start) ones. MAPF has many applications, including automated warehouses, autonomous vehicles, and video games and is being widely studied in the literature. Depending on the application, many variants of MAPF have been proposed~\cite{stern2019mapf} and numerous solutions have been already presented. One variant is the Anonymous MAPF (AMAPF). In AMAPF, the agents are interchangeable and each agent may be assigned to any goal, assuming that, in the end, all goal locations will be reached (in case the number of goal locations is smaller than or equal to the number of agents) or each agent will arrive to one goal location (otherwise). %In other words, the goal locations should be assigned to the agents and  the collision-free paths to them should be found.
This problem naturally arises in such environments where the tasks can be performed by any agent, e.g. identical robots carrying packages/inventory pods in automated warehouses.

In this work we aim to solve the AMAPF problem optimally w.r.t \emph{makespan} cost function, which is the arrival time of the last agent. In other words, our task is to find a solution where the last agent arrives at its goal location as early as possible. 
State-of-the-art optimal AMAPF solvers are reduction-based, i.e. the initial problem is reduced to another one and the latter is solved with an off-shelf solver. In the case of AMAPF, the common reduction is the following. Based on the input graph, another one is constructed. Then a maximum flow problem on this auxiliary graph, called a network, is formulated and solved. The latter can be interpreted as finding several paths (subject to certain constraints) on the reduced network. The major bottleneck here is that the size of the network is much larger, both in the number of vertices and edges, compared to the initial AMAPF graph, and, therefore, finding paths on it is burdensome. Moreover, the AMAPF reduction scheme, in general, assumes that numerous networks may be consecutively constructed (each one being larger than the previous one) and the search should be repeated.

To this end, we present an improved optimal AMAPF solver that follows the reduction-to-the-flow-problem approach while utilizing a novel search method to find paths on the (flow) networks. The crux of our search method is the concept of bulk states and implicit expansions. In brief, instead of generating and expanding numerous search states, we compress them into the bulks that form a sequence, exploiting the special structure of the underlying network, and explicitly store in the search tree only the certain representatives of those bulks (while implicitly reasoning about all other states in the bulk). On the theoretical side, we show that our search method, dubbed Bulk Search, is complete. On the practical side, we compare our improved AMAPF solver that utilizes Bulk Search with the state-of-the-art optimal AMAPF solver and show that our algorithm notably scales better to large maps (due to significantly lower number of expansions when finding the paths on the flow networks) and outperforms the competitor on all maps of the well-known MAPF benchmark from~\cite{stern2019mapf}.

\section{Related Works}
In a conventional MAPF formulation~\cite{stern2019mapf} a set of agents is given as well as the specification where each agent starts and the goal it should reach. Even when both the time is discretized into time steps and workspace is discretized into a graph (which are the two default assumptions in MAPF), obtaining an optimal solution w.r.t. one of the most widely-used objectives, e.g. the makespan or the sum-of-costs, is known to be NP-hard~\cite{yu2013planning}. Surprisingly, the AMAPF problem, which is a combined problem of both MAPF and goal assignment, can be optimally solved w.r.t. makespan (but not the sum-of-costs) in polynomial time~\cite{yu2013multi}. The seminal method, introduced in~\cite{yu2013multi}, is based on the reduction of AMAPF to a series of specific graph-search problems, i.e. the problems of finding a maximum flow~\cite{ford2015flows} on a graph of special structure (network) induced by the input MAPF graph. For the sum-of-costs objective an adaptation of the seminal MAPF solver, Conflict-Based Search (CBS)~\cite{sharon2015conflict} was suggested in~\cite{honig2018conflict}. Indeed, this algorithm is not polynomial. Suboptimal AMAPF was studied in~\cite{okumura2022solving}; several computationally efficient algorithms were proposed in this work which were empirically shown to provide high-quality solutions. However, no bound on sub-optimality was theoretically guaranteed. A variant of the AMAPF problem with some additional practically inspired assumptions, i.e., that the number of goals exceeds the number of agents and thus agents have to move to the new goals upon completing the current ones, was explored in~\cite{nguyen2017generalized} and solved using the Answer Set Programming (ASP). 

More involved variants of AMAPF were studied in~\cite{ma2016optimal,bartak2021fromclassical}. It was assumed that the agents are partitioned into the teams (colors) and each team is assigned a set of interchangeable targets (of the same color). In~\cite{ma2016optimal}, a combination of CBS and min-cost max-flow algorithm~\cite{ford2015flows}
was suggested to solve this Colored MAPF problem. \citeauthor{bartak2021fromclassical}~\shortcite{bartak2021fromclassical} proposed several solvers that utilize reduction to SAT. Indeed, AMAPF can be viewed as a special instantiation of the Colored MAPF problem (i.e. the one when there exists only a single team of agents of the same color as all the goals).

Among the other problems that are closely related to AMAPF, one can name Lifelong MAPF (LMAPF)~\cite{li2021lifelong} and Multi-agent Pickup and Delivery (MAPD)~\cite{ma2017lifelong}. These MAPF variants assume that the agents continuously operate in the environment reaching the specified goals (associated with certain pickup-and-delivery tasks in the case of MAPD). However, the assignments of goals (tasks) to agents is commonly assumed to be realized by an external procedure and, thus, the assignment sub-problem is not typically considered as part of the LMAPF/MAPD problem. Still, there are papers that consider a combined problem~\cite{chen2021integrated,xu2022multi}.

Finally, a body of works studies AMAPF in continuous domains, i.e. not assuming that the agents are confined to a given graph but are rather allowed to freely move in the (geometric) workspace~\cite{adler2015efficient,solovey2016hardness}.

\section{Problem Statement}
We follow a classical approach~\cite{stern2019mapf} to define the problem under investigation -- AMAPF. We consider a graph $G=(V,E)$, whose vertices correspond to the locations in the environment and edges -- to the transitions between them. $k$ agents are confined to this graph, i.e., initially each agent occupies a (distinct) vertex -- $s_i$, the start vertex, and at each time step of the discretized timeline, it can either wait in its current vertex or move to an adjacent one. The duration of both types of actions (move or wait) is $1$ time step. $k$ goal vertices, $g_1, ..., g_k$, are also distinguished. It is assumed that any agent can reach any goal, i.e., there is no pre-defined assignment of agents to the goals.

A plan for an agent, $\pi(s, g)$, is a sequence of (move/wait) actions, s.t. it begins at vertex $s$ and ends at vertex $g$; each action in the plan starts where the the previous one ends. The cost of the plan is the time step by which $g$ is reached. Two plans are said to contain a vertex (similarly, an edge) conflict if the agents following them occupy the same vertex (use the same edge) at the same time step. 

The problem now is to find a set of plans $\Pi=\{\pi_1, ..., \pi_k\}$, s.t. (1) each pair of plans is conflict-free and (2) all goals are reached. Essentially, this problem is a combination of the assignment problem, where one needs to decide which agent goes to which goal, and the (multi-agent) pathfinding problem, where one needs to construct a set of conflict-free plans.

We consider the following cost objective: $makespan(\Pi)=\max_{i\in{1,...,k}}(cost(\pi_i))$, where $cost(\pi)$ is the cost of the individual plan (i.e. the earliest time step when the agent reaches a goal vertex and never moves away). In this work, we are interested in obtaining makespan-optimal solutions of the problem at hand (AMAPF).

\section{Background}
\subsection{Network Flow}
Generally, network flow problem might come in different flavors; see~\cite{ahuja1995network} for an overview. Here we focus on a specific variant of the problem needed for solving AMAPF problems.

A network is a tuple $N = (G,cap,s,g)$, where $G = (V,E)$ is a directed graph, $cap : E \rightarrow Z^+$ is the mapping defining the capacities of the edges, $s \in V$ is the source vertex, and $g \in V$ is the sink vertex. For vertex $v \in V$, let $\sigma^+(v)$ (resp. $\sigma^-(v)$) denote the set of edges of $G$ going to (resp. leaving) $v$. A feasible $s,g$-flow on the network is mapping $f : E \rightarrow Z^+$ that satisfies three types of constraints: edge capacity constraints:
\begin{equation} 
\forall e \in E, f(e) \leq cap(e),
\end{equation}
the flow conservation constraints at non-terminal vertices:
\begin{equation}
\label{eq:conserv1}
    \forall v \in V \setminus \{s,g\}, \sum_{e\in\sigma^+(v)}f(e) - \sum_{e\in\sigma^-(v)}f(e) = 0,
\end{equation}
and the flow conservation constraints at terminal vertices:
\begin{equation}
\label{eq:conserv2}
    F(f) = \sum_{e\in\sigma^-(s)}f(e) = \sum_{e\in\sigma^+(g)}f(e).
\end{equation}

The quantity $F(f)$ is called the \textit{value} of the flow $f$. Another interpretation of the flow is that the flow is a set of $s$-$g$ paths (possibly overlapping or even duplicating), where each path caries a unit of flow from $s$ to $g$, such that the sum of units passing through any edge does not exceed its capacity.

The standard single-commodity maximum flow problem asks the following question: Given a network $N$, what is the maximum $F(f)$ that can be pushed through the network? Alternatively, find a set of $s$-$g$ paths that carry the maximum units of flow through the network.

\subsection{From AMAPF to Network Flow}
In~\cite{yu2013planning}, the authors reduced the $T$-steps AMAPF problem, i.e. the one that allows any agent to do at most $T$ actions, to a maximum flow (MF) problem. Specifically, it was proven that a $T$-steps AMAPF problem has a solution \textit{iff} the reduced MF problem has a flow equal to the number of agents. The makespan for an AMAPF instance can therefore be found by finding the smallest $T$ such that $T$-steps AMAPF instance has a solution. We now explain the suggested reduction with a slight modification suggested by~\cite{liu2019task} that simplifies it.

\begin{figure}[!t]
    \centering
\includegraphics[width=1\columnwidth]{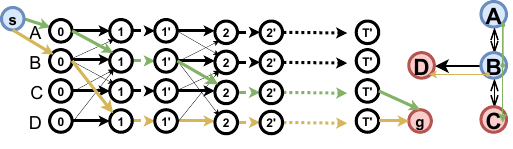}
    \caption{Example of the flow network (left) for a $T$-steps AMAPF instance (right). 
    Each line in the flow network represents the copies of a single vertex in the original AMAPF graph.
    The diagonal, solid horizontal, and dashed horizontal edges in the network denote \textit{move}, \textit{wait}, and \textit{restriction} edges, respectively.
    In this example, the AMAPF instance has two start vertices $A,B$ and two goal vertices $C,D$, so the source node $s$ in the network is connected to the nodes $(A,0),(B,0)$ and the sink $g$ is connected to $(C,T'),(D,T')$. The example also shows the matching between the plans for the agents and the $s$-$g$ paths in the network (green and yellow edges).}
    \label{fig:mf-network}
\end{figure}

Consider a $T$-steps AMAPF instance with the graph $G=(V,E)$. We first create $2T+1$ copies of $V$ and mark them as follows: $0,1,1',2,2',...,T'$; see Fig.~\ref{fig:mf-network}. Hereinafter, we will use the term vertices to denote the elements of the original AMAPF graph and the term nodes to denote the elements of the constructed network. We will also call copies $t'$ (with apostrophe) and copies $t$ for $t=0,1,..T$ as outer and inner copies, respectively. The copied vertices of the original graph form the nodes of the network. Each node is identified by $(v,h)$, where $h$ stands for the copy, alternatively referred to as height. 
% Higher node means greater copy, and $h'$ is higher than $h$. 
Indeed, the node $(v,h)$ (as well as $(v, h')$) corresponds to the state of an agent located in vertex $v$ at time step $h$.

Then, for each edge $e(u,v)$ in the original graph, we connect the nodes $(u,h')$ and $(v,h+1)$ for $h>0$, and also connect $(u,0)$ and $(v,1)$. These edges correspond to the move actions, and we call them the \textit{move} edges. Then, we add the \textit{wait} edges that connect the nodes $(u,h')$ and $(u,h+1)$ for $h>0$ and the nodes $(u,0)$ and $(u,1)$. Additionally, we add the edges between the nodes $(u,h)$ and the $(u,h')$. These edges do not denote any action but are added to forbid any two $s$-$g$ paths in the network from sharing any node and, therefore, to avoid node-collisions. We call them the \textit{restriction} edges. Finally, we add the source $s$ and sink $g$ nodes and connect $s$ to all nodes $(u,0)$ where $u$ is a start vertex, and connect $g$ to nodes $(v,T')$ where $v$ is a goal vertex. The capacity of any edge is fixed to be 1.

Hereinafter, whenever a path in a network is mentioned we will mean the $s$-$g$ path. The matching between the AMAPF and MF is now straightforward. Each plan for an agent can be matched to a path in the network by matching the agent actions to the \textit{move} or \textit{wait} edges and using the \textit{restriction} edges to connect between them. Similarly, a path in the network is matched to a plan for an agent where the \textit{move}/\textit{wait} edges are matched to move/wait actions. See Fig.~\ref{fig:mf-network} for a self-contained example. It was proven that there are no shared nodes in any two paths in the network (that form the solution to the MF problem), which infers that all matched plans have no node-collisions. An edge-collision may happen if two paths pass the edges $((u,h')\rightarrow(v,h+1))$ and $((v,h')\rightarrow(u,h+1))$ as these two different edges in the network refer to the same edge in the original graph. However, using the approach suggested in~\cite{liu2019task}, these collision can be eliminated in the following fashion. Instead of two conflicting agents moving to their next vertices, they swap plans and continue moving by the other's plan.
As a result, the AMAPF solution can be obtained by finding the maximum number of paths (maximum flow) in the described network.

\subsection{Solving Maximum Flow}

Ford-Fulkerson algorithm~\cite{ford1956maximal} was suggested in~\cite{yu2013planning} to solve the maximum flow problem. The algorithm is simple and easy to implement; its complexity on the reduced networks is $O(kEV)$ (formulated in~\cite{yu2013planning}), where $E$, $V$ stand for the number of nodes and edges in the original graph. 
Since, in our MF problem, all the edges have a capacity of one and the value of the flow is bounded by the number of agents (as each path is to be matched to an agent's plan), Ford-Fulkerson can be considered as one of the most efficient solvers for our case (see~\cite{cruz2023survey} for more details about Maximum Flow algorithms).

We refer the reader to the original paper for the detailed description of Ford-Fulkerson while briefly describing how the algorithm works specifically for the reduced network. In particular, the following two steps are sequentially repeated. First, we find a path $p$ from $s$ to $g$ in the network. Second, we reverse all edges of $p$. We keep repeating these two steps until no path can be found. The found paths form the maximum flow in the network (in our case, they form the plans of the agents in the original AMAPF problem)~\footnote{See the appendix for an illustrative example.}.

\section{Solving MF Efficiently On The Introduced Networks}\label{sec:solving_mcmf}

The maximum value of $T$ needed to solve an AMAPF problem can be up to $k+V-2$, as shown by~\cite{yu2013planning}. This implies that the network size may be quadratic in the number of vertices of the original graph.
Therefore, finding an $s$-$g$ path may become a bottleneck when solving AMAPF instances involving large input graphs.
To this end, we propose an algorithm that takes advantage of the specific structure of the reduced network and, as a result, is able to solve AMAPF problems much faster.

\begin{figure}[!t]
    \centering
\includegraphics[width=1\columnwidth]{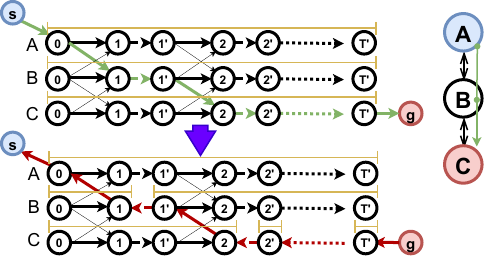}
    \caption{Example showing connected-sequences on the network. Yellow crossbars denote the connected-sequences on each vertex. Initially, we have the connected-sequences as shown in the upper figure. After a path (green one) is found and its edges are reversed, the connected-sequences are divided as shown in the lower figure.}
    \label{fig:mf-networks-with-connected-sequences}
\end{figure}

First, we present some definitions to use in the algorithm description.
In our algorithm, the search state corresponds to the network node. We will use $n(v,h)$ to denote the search state. Recall that the network node is defined by the vertex in original graph, $v$, and the height (copy) $h$ (a higher node means a higher copy, and $h'$ is higher than $h$).
Let us define a \textit{connected-sequence}
as a sequence of nodes with the same vertex in which we can achieve the last node from the first node using only \textit{wait} and \textit{restriction} not-reversed edges, and it cannot be extended in either side (i.e., it has the maximum length). An example is shown in Fig.~\ref{fig:mf-networks-with-connected-sequences}. Initially, for each vertex $v$ of the original graph, we have only one connected-sequence $(v,[0,T'])$ i.e. the one that starts with the node with height $0$ and ends with height $T'$ (shown by yellow crossbars). After a path (green path) is found, its edges are reversed (red edges in the lower graph). As a result, some connected-sequences disconnect which leads to several connected-sequences at the same vertex (see the lower network). %Next we describe the details of the algorithm.

\paragraph{Idea} The suggested algorithm is a graph traversal algorithm where the order of the search states in OPEN (the set of nodes that are the candidates to be expanded at the next iteration of the algorithm) is determined by their heights, i.e., the states with the lower heights are expanded first. 
The crucial idea of the algorithm is to expand states in \textit{bulks} while searching. 
That is, instead of expanding one search state, i.e. generating its successors and marking it as visited (adding it to CLOSED), in each search iteration, we (implicitly) expand a bulk of states at once. 
Such bulk expansion can be effectively implemented using the introduced notion of the connected sequence, resulting in the reduction of the expansions number, time and memory compared to expanding states individually.
Next, we describe how we can form the bulks of states and how the successors can be found compactly and fast.

We note that a similar but more specific concept was used in ~\cite{phillips2011sipp} and~\cite{gonzalez2012using} to implicitly compress and expand the states generated by the wait actions of agents~\footnote{The mentioned algorithms were originally tailored to graph-based pathfinding in the presence of dynamic obstacles where each vertex of the graph had to be annotated with the (safe) time intervals and the search utilized implicit move-then-wait actions. Our algorithm on the other hand handles explicit graphs, and the notion of connected-sequence in general is not obligated to relate to the time dimension, as will be shown later in the paper.}.
Let us assume that while traversing the graph naively by single nodes, we are to expand a state $(v,h)$ located in a connected-sequence $(v,[h_{min},h_{max}])$.
We can note that the nodes with the higher heights inside the connected-sequence (i.e. the nodes $(v,x):x\in[h+1,h_{max}]$) are all achievable from $(v,h)$ via the \textit{wait} and \textit{restriction} edges. 
Hence, the idea is to directly (and implicitly) generate all of these states ($(v,x):x\in[h+1,h_{max}]$), form an \textit{implicit} bulk consisting of these states (including the originally picked-up node $(v,h)$), and expand them all at once (instead of expanding only $(v,h)$). We will refer to the described mechanism of generating the sequential successors of states that uses only  \textit{wait} and \textit{restriction} edges as \textit{straightforward} generation. Note that it is enough to store the vertex and the height bounds of the bulk to define it. So, technically, when forming a bulk (for future expansion), we only need to find the last \textit{straightforwardly} achievable state (i.e. the highest state in the connected-sequence).

When the bulk is ready, it is expanded in the following fashion. First, let us assume the naive expansion of a bulk when, for every node that resides in it, we generate all of its immediate successors. Now observe that, as a result of such procedure, we are likely to have numerous successors that are characterized by the same graph vertex and different heights. Moreover, many of these successors may belong to the same connected-sequence. Thus, instead of explicitly generating them, we generate only the ones with lower heights in their sequences. The other search states from these sequences will be \emph{straightforwardly} generated later on (i.e., when the search state with the lowest copy is picked for expansion and its bulk is formed as described above).

In other words, to expand a bulk of states $(v,[h_l,h_u])$, we do the following. We iterate over all connected-sequences in neighbor vertices of $v$, in which we can achieve at least one node (from a node $(v,x):x\in[h_l, h_u]$). Then, we only generate the accessible node with the minimum height in each of these connected-sequences. The periodic structure of the reduced networks allows us to quickly find the node with the minimal height (as we will show later). 
As the number of connected-sequences is much less than that of individual nodes, it leads to a high reduction in the number of generated states. We call the algorithm that utilizes the described concepts Bulk Search (BS). Next, we describe the details of the implementation.

\begin{algorithm}[t!]
\small
\caption{Bulk Search}
\label{alg:main}
\begin{algorithmic}[1]
\renewcommand{\algorithmicrequire}{\textbf{Input:}}
\renewcommand{\algorithmicensure}{\textbf{Output:}}
    \REQUIRE Network $N(G,cap,s,g)$
    \ENSURE Path from $s$ to $g$ if exists
    \STATE OPEN $\leftarrow \phi$, CLOSED $\leftarrow \phi$
    \STATE insert $s\rightarrow$ OPEN
    \WHILE{OPEN $\neq \phi$}
        \STATE remove $n(v,h)$ with the minimum $h$ 
        \color{black}from OPEN
        \IF{$n = g$}
            \RETURN path from $s$ to $n$
        \ENDIF
        \IF{$n$ $\in$ CLOSED}
            \STATE continue 
        \ENDIF
        % \color{red}
        \STATE $x(v,h')\gets$ the state from CLOSED in the same connected-sequence of $n$ and minimum height
        \IF{$x$ exists \AND $h'<=h$}
            \STATE continue
        \ENDIF
        \color{black}
        \STATE $succ \leftarrow$ $getSuccessors(n)$
        \color{black}
        \FOR{$n'(u,h')$ in $succ$}
        \IF{$n'$ $\in$ OPEN $\cup$ CLOSED}
        \STATE continue
        \ENDIF
        % \color{red}
        \STATE $x(u,h'')\gets$ the state from OPEN $\cup$ CLOSED in the same connected-sequence of $n'$ and minimum height
        \IF{$x$ exists \AND $h''<=h'$}
            \STATE continue
        \ENDIF
        \color{black}
        \STATE insert $n'$ $\rightarrow$ OPEN
        \ENDFOR
    \ENDWHILE
    \RETURN no answer
% }
\end{algorithmic}
\end{algorithm}

\begin{algorithm}[t!]
\caption{Generating successors}
\label{alg:getsucc}
\small
\renewcommand{\algorithmicrequire}{\textbf{Input:}}
\renewcommand{\algorithmicensure}{\textbf{Output:}}

\begin{algorithmic}[1]
    \REQUIRE Network $N(G,cap,s,g)$, Node $n(v,h)$
    \ENSURE The successors of $n(v,h)$
   \IF{$n = s$}
   \RETURN all neighbor nodes of $n$ in $G$
   \ENDIF
   \STATE $succ \leftarrow \phi$
   \STATE $[h_{min}, h_{max}] \leftarrow$ height bounds of the connected-sequence where $n$ is located
   \IF{$h = h_{min}$}
        \STATE insert all neighbor nodes of $n$ in $G$ $\rightarrow succ$
   \ENDIF
   \STATE insert all neighbor nodes of node $(v, h_{max})$ in $G$ $\rightarrow succ$
   \FOR{each connected-sequence $cs(u,[h_l,h_u])$ with vertex $u$ is a neighbor of $v$}
   \STATE $c'_{from}\gets$ the height of the minimum outer copy in $[max(h_{min}+1, h),h_{max}-1]$
   \STATE $c_{to}\gets$ the height of the minimum inner copy in $[h_l,h_u]$
        \IF{ $c'_{from}+1<=h_u$ \AND $c_{to}-1<=h_{max}$
        % $\exists$ node $(u, i)$ from $cs$, is accessible from a node $(v,j):j\in[max(h_{min}+1, h),h_{max}-1]$ 
        }
            \STATE $c_{min}\gets max(c_{to},c'_{from}+1)$
            \STATE insert $(u,c_{min}) \rightarrow succ$
        \ENDIF
   \ENDFOR
   \RETURN $succ$
   
\end{algorithmic}
\end{algorithm}

\paragraph{Details} Algorithm~\ref{alg:main} shows the pseudo-code of a graph traversal algorithm with our modifications. First, we order the states inside the search set (OPEN) by their height, i.e., we always choose the state with the minimum height for the expansion. This will help us reduce the number of expansions as will be shown later. The second change is that whenever we need to check whether a chosen state was expanded before, we additionally check if the state can be \textit{straightforwardly} generated from the previously opened states. In this case, we do not need to expand it, as we assume that it was implicitly expanded before and its successors were already generated and inserted into the search set.
This can be done by checking if any state in the same connected-sequence and lower height was expanded before (i.e. stored in the CLOSED set) (lines 11-14). This check should also be done when inserting new states into the OPEN set (lines 20-23).

The third change is how we generate the successors of all states (the taken one from OPEN along with its \textit{straightforwardly} generated states) fast. The pseudo-code is presented in Algorithm~\ref{alg:getsucc}. 
Firstly, (lines 1-3), if the node is the source node $s$, we have only one node as input (i.e. no implicit states), and, thus, we need to only generate neighboring successors (i.e. connected nodes in the network) and return them. Otherwise, let the input node $n(v,h)$ (which is not $s$ or $g$) located in the connected-sequence $(v, [h_{min}, h_{max}])$, then we should generate the successors of all states $n(v,x):x\in [h,h_{max}]$. This can be done as follows. If the state $(v, h)$ is located at the beginning of the connected-sequence (i.e. $h=h_{min}$), this state may have reversed edges, so we always generate all neighboring nodes of this state in the network. The same thing is applied on $(v,h_{max})$ where we also generate all its neighboring nodes in the network.
Other nodes (i.e. nodes $(v,x):x\in[max(h_{min}+1, h),h_{max}-1]$) have only \textit{move} edges to connect to nodes in other connected-sequences, so we can do the following (lines 10-17) to generate their successors. We iterate over all connected-sequences $cs$ in neighbor vertices of $v$. We then check whether we can achieve at least one node in $cs$ (from nodes $(v,x):x\in[max(h_{min}+1, h),h_{max}-1]$). As shown in lines 11-13, this can be done by checking whether there is at least one \textit{move} edge which starts from an outer copy in $[max(h_{min}+1, h),h_{max}-1]$ and ends at inner copy in $cs$. If so, we generate the node with the minimum accessible height in $cs$ and add it to the successors set (lines 14-15).

As a result, any achievable node beginning from the source node is either explicitly or implicitly (from a node with lower height in the same connected-sequence) expanded. Therefore, we can immediately state the following theorem.

\begin{theorem}
BS is a complete algorithm.
\end{theorem}

\paragraph{Theoretical analysis}
The search states inside the search set are sorted according to the height. This helps to reduce the number of expansions as the lower-height states implicitly expand the higher-height states in the same connected-sequence, but the opposite is not applicable.
In theory, multiple single states in one connected-sequence may be expanded individually, e.g., if they have been opened in the descending-height way. This is possible even if we order the states according to the heights as we have reversed edges which generates states with a lower height. However, in practice, only few states are expanded in each connected-sequence, and, thus, the algorithm's performance mainly depends on the number of connected-sequences. Fortunately, the number of connected-sequences is much smaller than the size of the network. Initially, we have a number of connected-sequences equals $V$, the number of original graph vertices. After each path is found and reversed, $T$ new reversed edges, and therefore, $T$ new connected-sequences appear. As a result, in the whole search for all $k$ agents, the number of the appearing connected-sequences equals $\sum_{i=1}^{i=k}{|V|+T(i-1)}=k|V|+Tk(k-1)/2$. 
% The theoretical number of single nodes in the network for all agents is $k|V|T$. 
Therefore, we have a theoretical reduction in the number of nodes (comparing with $k|V|T$, the number of nodes without compressing in connected-sequences) equals $min(|V|/k, T/2)$ (i.e. $(k|V|+Tk(k-1)/2)*min(|V|/k, T/2) <= k|V|T$~\footnote{See the appendix for details.}). This reduction is significantly high, which allows us to obtain the fastest full success (to our knowledge) in optimally solving all public MAPF benchmarks for a MAPF-family problem, as will be shown in the next section.

\begin{figure}[t!]
    \centering
\includegraphics[width=0.7\columnwidth]{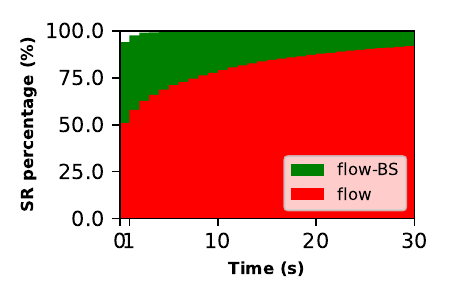}
    \caption{The (normalized) number of instances solved by a certain time cap.
    % while searching for the optimal makespan in AMAPF/AMAPFD problems, for both ITRTV-MF and MCMF+SIPP algorithms.
    }
    \label{fig:results-sr-hist}
\end{figure}

\begin{table}[t!]
{
\small
\centering
\setlength\tabcolsep{3.5pt}
% \begin{tabular}{m{3cm}|m{1cm}|m{1cm}|m{1cm}|m{2cm}}
\begin{tabular}{c|c|c|c|c}
\multirow{2}{*}{Map} & \multirow{2}{*}{Width} & \multirow{2}{*}{Height} & \multicolumn{2}{c}{Algorithm}\\
\cline{4-5}
&&&flow & flow-BS\\
\hline
empty-8-8 & 8 & 8 & \textbf{100}\% & \textbf{100}\%\\
empty-16-16 & 16 & 16 & \textbf{100}\% & \textbf{100}\%\\
maze-32-32-2 & 32 & 32 & \textbf{100}\% & \textbf{100}\%\\
room-32-32-4 & 32 & 32 & \textbf{100}\% & \textbf{100}\%\\
maze-32-32-4 & 32 & 32 & \textbf{100}\% & \textbf{100}\%\\
random-32-32-20 & 32 & 32 & \textbf{100}\% & \textbf{100}\%\\
random-32-32-10 & 32 & 32 & \textbf{100}\% & \textbf{100}\%\\
empty-32-32 & 32 & 32 & \textbf{100}\% & \textbf{100}\%\\
empty-48-48 & 48 & 48 & \textbf{100}\% & \textbf{100}\%\\
den312d & 65 & 81 & \textbf{100}\% & \textbf{100}\%\\
room-64-64-8 & 64 & 64 & \textbf{100}\% & \textbf{100}\%\\
random-64-64-20 & 64 & 64 & \textbf{100}\% & \textbf{100}\%\\
room-64-64-16 & 64 & 64 & \textbf{100}\% & \textbf{100}\%\\
random-64-64-10 & 64 & 64 & \textbf{100}\% & \textbf{100}\%\\
warehouse-10-20-10-2-1 & 161 & 63 & \textbf{100}\% & \textbf{100}\%\\
ht\_chantry & 162 & 141 & \textbf{100}\% & \textbf{100}\%\\
maze-128-128-1 & 128 & 128 & 3\% & \textbf{100}\%\\
ht\_mansion\_n & 133 & 270 & 96\% & \textbf{100}\%\\
warehouse-10-20-10-2-2 & 170 & 84 & \textbf{100}\% & \textbf{100}\%\\
lt\_gallowstemplar\_n & 251 & 180 & 82\% & \textbf{100}\%\\
maze-128-128-2 & 128 & 128 & 4\% & \textbf{100}\%\\
ost003d & 194 & 194 & 57\% & \textbf{100}\%\\
lak303d & 194 & 194 & 44\% & \textbf{100}\%\\
maze-128-128-10 & 128 & 128 & 17\% & \textbf{100}\%\\
warehouse-20-40-10-2-1 & 321 & 123 & 91\% & \textbf{100}\%\\
den520d & 256 & 257 & 16\% & \textbf{100}\%\\
w\_woundedcoast & 642 & 578 & 1\% & \textbf{100}\%\\
warehouse-20-40-10-2-2 & 340 & 164 & 8\% & \textbf{100}\%\\
brc202d & 530 & 481 & 1\% & \textbf{100}\%\\
Paris\_1\_256 & 256 & 256 & 5\% & \textbf{100}\%\\
Berlin\_1\_256 & 256 & 256 & 6\% & \textbf{100}\%\\
Boston\_0\_256 & 256 & 256 & 4\% & \textbf{100}\%\\
orz900d & 1491 & 656 & 0\% & \textbf{100}\%\\

% \hline
\end{tabular}
}
    \caption{Success rates of \textit{flow-BS} and \textit{flow} solvers with a timeout of 30 seconds.}
    \label{table:sr-new-results}
% \end{center}
\end{table}

\section{Experimental Evaluation}

\begin{figure*}[t!]
    \centering
\includegraphics[width=1\textwidth]{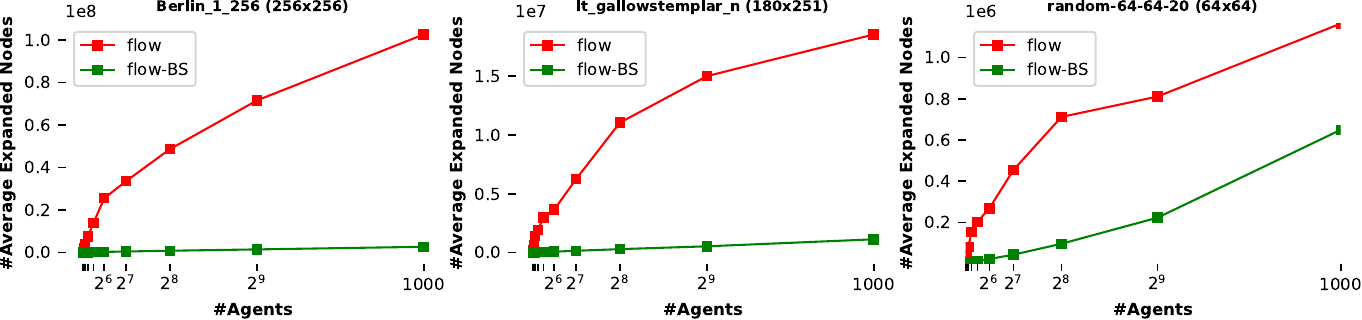}
    \caption{The number of expanded nodes with a different number of agents on different maps.
    % while searching for the optimal makespan in AMAPF/AMAPFD problems, for both ITRTV-MF and MCMF+SIPP algorithms.
    }
    \label{fig:results-expanded_nodes}
\end{figure*}

\begin{figure*}[h!]
    \centering
\includegraphics[width=1\textwidth]{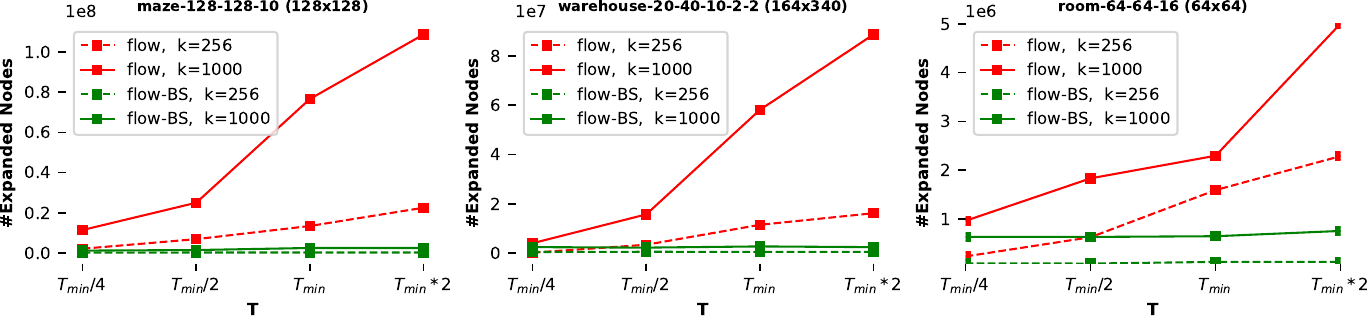}
    \caption{The figure shows (for a specific map and a specific number of agents) how the number of expansions changes with the increasing values of $T$. Here $T_{min}$ denotes the optimal makespan.
    }
    \label{fig:results-expanded_nodes-t}
\end{figure*}

We have implemented the improved AMAPF solver in C++\footnote{https://github.com/PathPlanning/AMAPF-MF-BS} and compared it with the state-of-the-art AMAPF solver that does not use the introduced Bulk Search to solve MF but rather utilizes the standard Ford-Fulkerson as suggested in~\cite{yu2013planning}. The code of the competitor was taken from public repository~\footnote{https://github.com/Kei18/tswap} that accompanied the paper~\cite{okumura2022solving}.
We kept all the optimization techniques designed by the code authors.
We will denote these two solvers as \textit{flow-BS} (ours) and \textit{flow} (state of the art). The experiments were conducted on a PC with Intel Core i7-10700F CPU @ 2.90GHz $\times$ 16 and 32Gb of RAM.

The MAPF maps and instances were taken from the publicly available MAPF benchmark~\cite{stern2019mapf}. We use all 33 maps available in this benchmark and all of the 25 random scenarios. Each scenario on each map (except some small ones) contains 1,000 pairs of start-goal positions. To test a solver on a scenario, we run it with the first 1, 2, 4, 8, 16, 32, 64, 128, 256, 512, and 1,000 pairs of start-goals sequentially. Whenever the solver fails to solve a problem under a time limit of 30 seconds, we terminate testing on this scenario and move to the next one. 

In the first experiment, we have used a precise estimator of $T$ suggested by~\cite{okumura2022solving} (which solves bottleneck assignment problem~\cite{gross1959bottleneck}) to estimate the lower bound of the makespan. As this estimator takes non-negligible time (up to 10 seconds) and both algorithms use it, we have not accounted for its runtime.

For each map, we have collected the total number of success instances over all scenarios. Table~\ref{table:sr-new-results} summarizes the results. As can be seen, our solver is able to solve all instances in all maps under 30s. However, this is not the case for \textit{flow} solver. The latter searches the network node-by-node and, therefore, is able to solve the test instances only on the small maps. When the maps are large (like city maps that are $256 \times 256$) or the makespan is high (like in the maze maps), it is often unable to produce a solution under the imposed time limit. Fig.~\ref{fig:results-sr-hist} shows the success rate for all instances in all maps second by second. In fact, \textit{flow-BS} is able to solve the hardest instance (\verb|orz900d| map, scenario 20 with 256 agents) in 17s. On the other hand, \textit{flow} has shown a very slight increase of SR after solving the easy instances (~75\% of instances) around the 8th second.

In the second experiment, we have investigated how the number of agents affects the performance. For this purpose, we selected three maps of different topology and size and plotted the average number of the expanded nodes against the number of agents. The results (plotted in Fig.~\ref{fig:results-expanded_nodes}) show that our algorithm expands much fewer nodes than the standard algorithm (as expected). 
% The graph of \textit{flow-BS} is consistent with the theoretical formula of the number of connected-sequences, and this again proves the dependence of our algorithm on the number of connected-sequences in the network and not the network size as in \textit{flow}.
It is worth noting that the relation between the runtime and the number of expansions is not fixed but increases for $flow$, too. This is because a single memory-access/read/write operation takes longer when the amount of the stored data increases.

So far, we have assumed that we have an estimator which can identify an exact bound ($T$) of the makespan. However, this may be not the case, so the search may be repeated several times until finding the optimal solution. 
Therefore, we have also conducted experiments to show the practical performance of both solvers for different $T$ when we fix the map and the number of agents. The tests are designed as follows. For each one of the three fixed maps, \verb|warehouse-20-40-10-2-2|, \verb|maze-128-128-10|, \verb|room-64-64-16|, and for a number of agents $\in \{256, 1000\}$, we run the solvers on networks with a maximum height equals one of the values $\{T_{min}/4, T_{min}/2, T_{min}, T_{min}*2\}$, where $T_{min}$ is the optimal makespan. Again, we have recorded the average number of expanded nodes over all scenarios (see Fig.~\ref{fig:results-expanded_nodes-t}). Obviously, the number of nodes expanded by \textit{flow} significantly increases with the value of $T$, while \textit{flow-BS} does not demonstrate such growth. It means that our solver is especially useful when one cannot estimate the optimal makespan accurately before actually solving an AMAPF problem instance.

\section{Conclusion}
In this paper, we have revisited the reduction-based approach to optimally solving the Anonymous MAPF problem, when the latter is reduced to a search problem on an auxiliary graph of a special structure. We have suggested an improved AMAPF solver that is based on a specific search algorithm tailored to find paths on the auxiliary graphs exploiting their specific topology. We have shown that our improved AMAPF solver significantly outperforms the state-of-the-art one on a large variety of setups, leveraging its better scalability to the size of the input graph. Next, we plan to extend the proposed search technique to support costs (e.g. Min-Cost-Maximum-Flow) to solve other MAPF problems.

\addtolength{\textheight}{0cm}   % This command serves to balance the column lengths

\section{Acknowledgments}
This work was partially supported by the Analytical Center for the Government of the Russian Federation in accordance with the subsidy agreement (agreement identifier 000000D730321P5Q0002; grant  No. 70-2021-00138).

\bibliography{aaai24}
\clearpage
\newpage
\input{appendix}
\end{document}

%% file: appendix.tex
\appendixpage
\appendix
\begin{figure*}[!t]
\centering
\includegraphics[width=\textwidth]{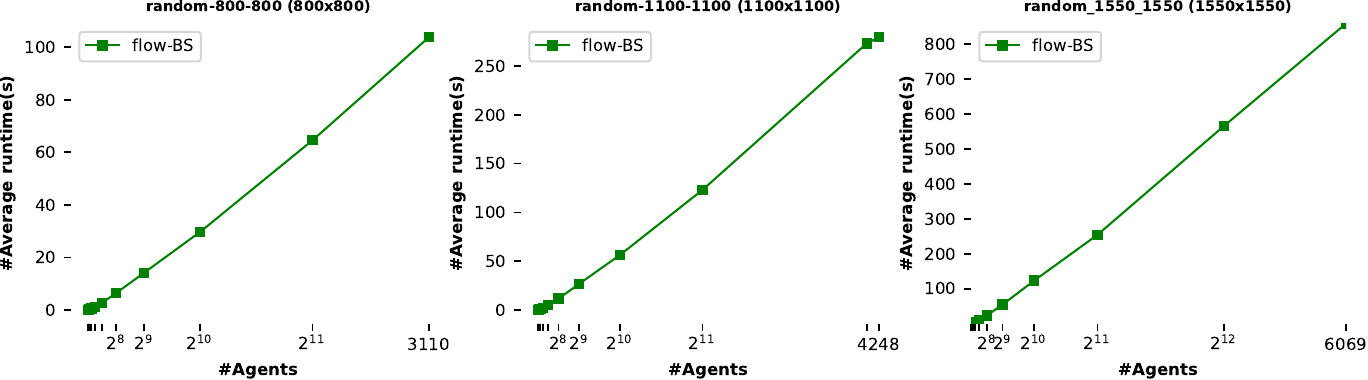}
\caption{Runtime of algorithm \textit{flow-BS} in large maps.}
\label{fig:big-maps}
\end{figure*}

\begin{figure*}[!t]
    \centering
\includegraphics[width=1\textwidth]{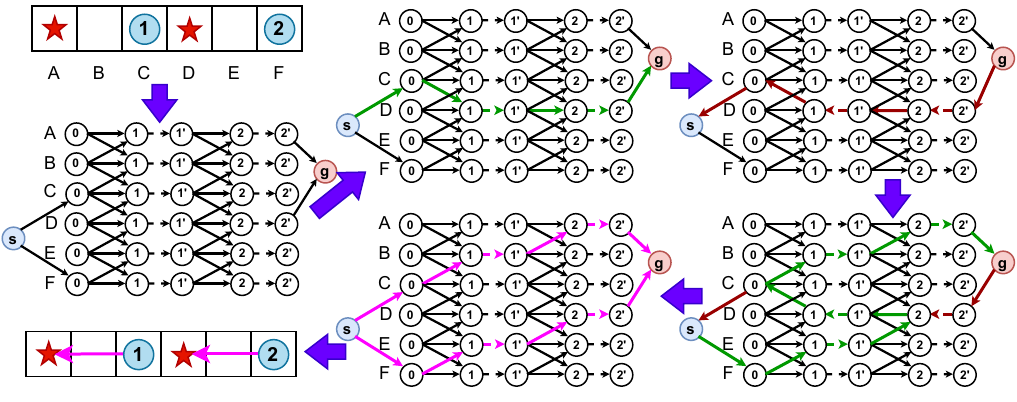}
    \caption{Example shows how the Maximum Flow algorithm works on the reduced-from-AMAPF networks step by step, and finally how the solution can be converted back to AMAPF solution. In the first image (in the upper left corner), shown the original AMAPF problem where we have two agents at cells C and F need to go to goal cells at A and D. In the next image we show the matching reduced network with maximum timestep $T=2$, where we need to solve MF problem. Next, assume that the green path is found from the source node $s$ to sink node $g$ (note that there are many paths leading from source to sink, and any one can be chosen). This path currently matches with the AMAPF plan in which the agent (1) should go to the goal at cell D. Next, we add a one unit flow to the edges of the path (i.e. to edges $((s),(C,0)), ((C,0),(D,1)), ((D,1),(D,1')), ((D,1'),(D,2)),((D,2),(D,2')),((D,2'),(g))$), and then reverse these edges to be used in next steps. Next, another green path is found from $s$ to $g$ which passes from some previously reversed edges as illustrated in the graph. We again add one unit flow to the edges of the path which were not-reversed and remove the flow from the edges of the path which were reversed ($((D,1),(D,1')), ((D,1'),(D,2))$). Next, no new path can be found from $s$ to $g$, so we mark all edges with positive flows and form the AMAPF plan from them. The resulting plan is: For agent (1), move from C to B, then move from B to A. For agent (2), move from F to E, then move from E to D.}
    \label{fig:amapf-mf-amapf-solution-example}
\end{figure*}
\subsection{Solving Maximum Flow}{From Flow solution to AMAPF solution}
We remind that in our special reduced networks, Ford-Fulkerson algorithm is reduced to two repeating steps. First, we find a path $p$ from $s$ to $g$ in the network. Second, we reverse all edges of $p$. We keep repeating these two steps until no path can be found. The found paths form the maximum flow in the network (in our case, they form the plans of the agents in the original AMAPF problem). Specifically, the flow is formed as follows. When a path passes from not-reversed edges, then a one unit of flow is added to those edges. Oppositely, when a path passes from reversed edges, then the flow is removed from these edges. At the end, each edge with positive flow is considered from the solution and therefore from the plan of the agents (according to the matching mentioned in the paper: \textit{move, wait} and \textit{restrict} edges converted to move, wait and nothing actions by the agents. See Fig.~\ref{fig:amapf-mf-amapf-solution-example} for an example.

\subsection{Additional metrics} 
In the paper, we decided to compare the number of expansions to make the comparison as much as possible independent from the implementation of \textit{flow}
and the hardware.
For reference, we add Table~\ref{table:runtime} to show the runtime of one expansion in the tests on the Berlin map. Please note that the runtime is bigger with smaller number of agents because the makespan, and therefore the stored network, is bigger. This is especially pronounceable for \emph{flow} as it stores the the whole network explicitly. As a result, the same (or better) superiority of \textit{flow-BS} is noticed when comparing the runtime.\\
\begin{table}[bh!]
    \centering
    \resizebox{\columnwidth}{!}{
    \begin{tabular}{|c||c|c|c|c|c|c|c|c|c||c|c|}
    \hline
         \#\_agents& 1 & 2 & 4 & 8 & 16 & 32 & 64 & 128 & 256 & 512 & 1000\\
         \hline
         Flow& 16686 & 4318 & 309 & 4 & 2 & 1 & 1 & 1 & 1 & 1 & 1\\
         \hline
         Flow-BS&1 & $<1$ & $<1$ & $<1$ & $<1$ & $<1$ & $<1$ & $<1$ & $<1$ & $<1$ & $<1$\\
         \hline
    \end{tabular}}
    \caption{Runtime of one expansion of both algorithms \textit{flow} and \textit{flow-BS} in microseconds for Berlin\_1\_256 map.}
    \label{table:runtime}
\end{table}

\subsection{Additional Tests} We ran additional experiments involving larger map sizes and agents number (taken from the \textit{Iron Harvest} benchmark). The results are reported in Fig.~\ref{fig:big-maps}. They confirm that our solver scales linearly with the number of agents and the size of the map. This is consistent with the theoretical estimate of the number of connected-sequences stated in the paper (where here the first part of the estimate ($k|V|$) is the dominant part).

\subsection{Proof for the theoretical reduction}
We state that the reduction in number of search nodes using BS algorithms equals $min(|V|/k, T/2)$, which means that the number of original search nodes is more than the number of connected-sequences by at least $min(|V|/k, T/2)$.  
Originally, in basic algorithms, each agent (from $k$) searches the whole network nodes $|V|T$.
As a result, the total number of search nodes can be up to $k|V|T$. It was deduced in the paper that the number of connected-sequences (which our algorithm depends on) is bounded by the value $k|V|+Tk(k-1)/2$ for all agents. Formally: \\
\\
\noindent\textbf{Claim:}\\
\scriptsize
\begin{equation*}
% \begin{split}
k|V|T >= min((|V|/k, T/2) * (k|V| + Tk(k-1)/2)
% \end{split}
\end{equation*}

\normalsize

\noindent\textbf{Proof:}\\

\indent If $\scriptstyle |V|/k <= T/2$ then
\scriptsize
\begin{equation*}
\begin{split}
% \begin{align*}
    &min(|V|/k, T/2) * (k|V| + Tk(k-1)/2)\\
    &=(|V|/k) * (k|V| + Tk(k-1)/2) \\
    &= (|V|/k) * (k|V|) + T|V|(k-1)/2\\
    &<= (T/2) * (k|V|) + T|V|(k-1)/2 \\
    &= k|V|T/2 + (k-1)|V|T/2 <=k|V|T
% \end{align*}
\end{split}
\end{equation*}
\normalsize

If  $\scriptstyle T/2 <= |V|/k$ then
\scriptsize
\begin{equation*}
\begin{split}
        &min(|V|/k, T/2) * (k|V| + Tk(k-1)/2)\\
        &= (T/2)*(k|V| + Tk(k-1)/2)\\
        &=k|V|T/2 + (T/2) * Tk(k-1)/2 \\
        &<=k|V|T/2 + (|V|/k) * Tk(k-1)/2\\
        &=k|V|T/2 + (k-1)|V|T/2 <=  k|V|T
\end{split}
\end{equation*}
% \hline